\documentclass[runningheads]{llncs}
\usepackage[T1]{fontenc}
\usepackage{pifont}   
\usepackage{multirow}  
\usepackage{graphicx,verbatim}
\usepackage{amsmath}
\usepackage{utfsym}
\usepackage{microtype}
\usepackage{marvosym}

\let\svthefootnote\thefootnote
\newcommand\freefootnote[1]{%
  \let\thefootnote\relax%
  \footnotetext{#1}%
  \let\thefootnote\svthefootnote%
}

\begin{document}
\title{Structured Reasoning Agentic Framework for Interpretable Critical View of Safety Assessment}
\titlerunning{Structured Reasoning Agentic Framework for Critical View of Safety}
%

\author{Qing Xu\inst{1}\textsuperscript{*} \and
Yuxiang Luo\inst{2}\textsuperscript{*} \and 
Zhen Chen\inst{2}\textsuperscript{(\Letter)}}

\authorrunning{Q. Xu et al.}

\institute{
School of Computer Science, University of Nottingham, UK \and
Department of Data Science and Artificial Intelligence,\\The Hong Kong Polytechnic University, Hong Kong SAR 
\\
\email{z.chen@polyu.edu.hk}
}

\maketitle              
\begin{abstract}
Surgical scene understanding is critical for computer-assisted intervention, yet laparoscopic cholecystectomy remains challenged by the complex anatomy of the hepatocystic triangle and the risk of bile duct injury. Existing methods for Critical View of Safety (CVS) assessment typically treat it as a holistic prediction task, mapping visual features directly to criterion-level labels. This black-box paradigm lacks explicit reasoning about anatomical relationships, limiting both interpretability and compositional generalization. To address this, we propose ReasonCVS, a structured reasoning agentic framework empowered by Vision-Language Models (VLMs) that decomposes CVS assessment into explicit, fine-grained anatomical verification. Specifically, we devise an Anatomical Scene Graph Abstraction (ASGA) that organizes anatomical entities and their spatial relationships into a structured representation. To operationalize this, we introduce a Rationale-Aware Reasoning Agent, powered by a Large Language Model (LLM) fine-tuned via rationale distillation. Functioning as a strict central decision-maker, it invokes VLM-driven Sub-criterion Verifier as a specialized perceptual tool to parse the graph and independently evaluate individual sub-criteria. Through calibrated soft reasoning, this agent synthesizes the tool-gathered distributed observations, yielding a final verdict alongside a traceable clinical rationale. Extensive experiments on the Endoscapes-CVS201 benchmark demonstrate that ReasonCVS achieves superior performance (68.1\% mAP) over state-of-the-art while providing interpretable, criterion-level explanations for reliable surgical assessment. \freefootnote{$*$ Equal contribution.} 



\end{abstract}
\section{Introduction}

Surgical scene understanding plays a crucial role in computer-assisted intervention and has received extensive attention in the research \cite{maier2017surgical,maier2022surgical,hashimoto2018artificial}. Particularly, laparoscopic cholecystectomy is challenged by the complex anatomy of the hepatocystic triangle and the need to accurately identify critical structures to prevent bile duct injury \cite{strasberg1995analysis,way2003causes}. These requirements have motivated a key safety task in the field, \textit{i.e.}, automated Critical View of Safety (CVS) assessment~\cite{strasberg2010rationale,brunt2020safe}.

Early works on CVS assessment formulated the problem as image-level classification, predicting criterion satisfaction directly from individual surgical frames \cite{mascagni2022artificial,twinanda2016endonet}, yet such frame-wise approaches prove insufficient to capture the anatomical complexity inherent to CVS evaluation. To incorporate spatial context, segmentation-based methods \cite{mascagni2022artificial,murali2023latent,madani2022artificial} introduced pixel-level anatomical representations for more fine-grained scene understanding. Graph-based approaches \cite{nwoye2022rendezvous,ozsoy20224d} have further explored structural modeling of surgical entities and their interactions. Moreover, recent multimodal approaches \cite{baby2025multi} have demonstrated the potential of vision-language pretraining, aligning surgical visual features with criterion-level textual descriptions to incorporate semantic knowledge beyond pure visual appearance.

Despite these advancements, existing methods \cite{mascagni2022artificial,murali2023latent,baby2025multi,rios2023cholec80} typically treat CVS assessment as a holistic prediction task, mapping visual features directly to criterion-level labels. This black-box paradigm lacks explicit reasoning about anatomical relationships, limiting both interpretability and compositional generalization. In fact, CVS evaluation inherently depends on the structural relationships among key anatomical entities in the hepatocystic triangle \cite{mascagni2022computer}. For example, confirming hepatocystic triangle clearance requires verifying that the cystic duct and cystic artery are the only two structures entering the gallbladder, demanding fine-grained relational reasoning rather than global appearance matching. Therefore, this strong structural dependency motivates us to decompose CVS assessment into fine-grained anatomical verification with a traceable rationale.
\begin{figure}[!t]
    \centering
    \includegraphics[width=1\linewidth]{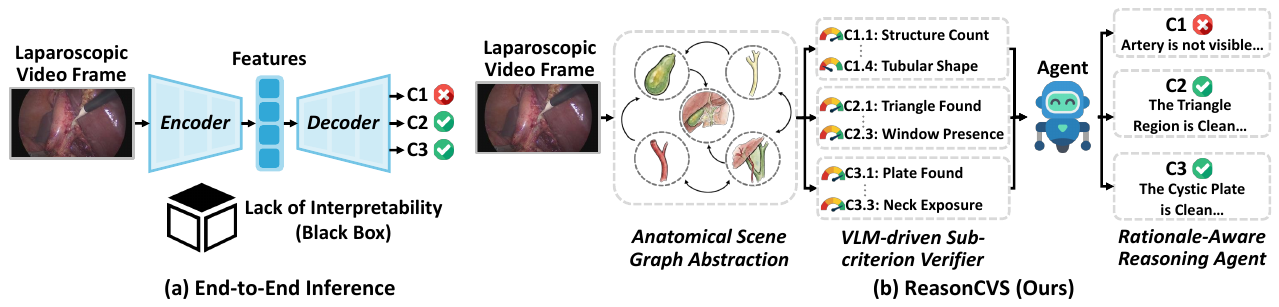}
    \caption{Comparison of existing methods and our ReasonCVS for CVS assessment. (a) Existing methods map visual features directly to criterion-level labels for holistic prediction. (b) Our ReasonCVS abstracts the surgical scene into an anatomical scene graph and invokes sub-criterion Verification for interpretable assessment.}
    \label{fig:placeholder}
\end{figure}

To address this, we propose \textbf{ReasonCVS}, a structured reasoning agentic framework empowered by Vision-Language Models (VLMs) that decomposes CVS assessment into explicit, fine-grained anatomical verification. Specifically, we first devise an Anatomical Scene Graph Abstraction (ASGA) that organizes anatomical entities and their spatial relationships into a structured representation. We then design the VLM-driven Sub-criterion Verifier to parse the graph and independently evaluate individual sub-criteria. A Rationale-Aware Reasoning Agent, fine-tuned via rationale distillation, serves as the central decision-maker that invokes these Verifications and synthesizes distributed observations through calibrated soft reasoning, yielding a final verdict alongside a traceable clinical rationale. To the best of our knowledge, ReasonCVS is the first framework to address CVS assessment through structured reasoning. Experiments demonstrate that ReasonCVS achieves superior performance over state-of-the-art methods.

\begin{figure}[!t]
    \centering
    \includegraphics[width=0.95\linewidth]{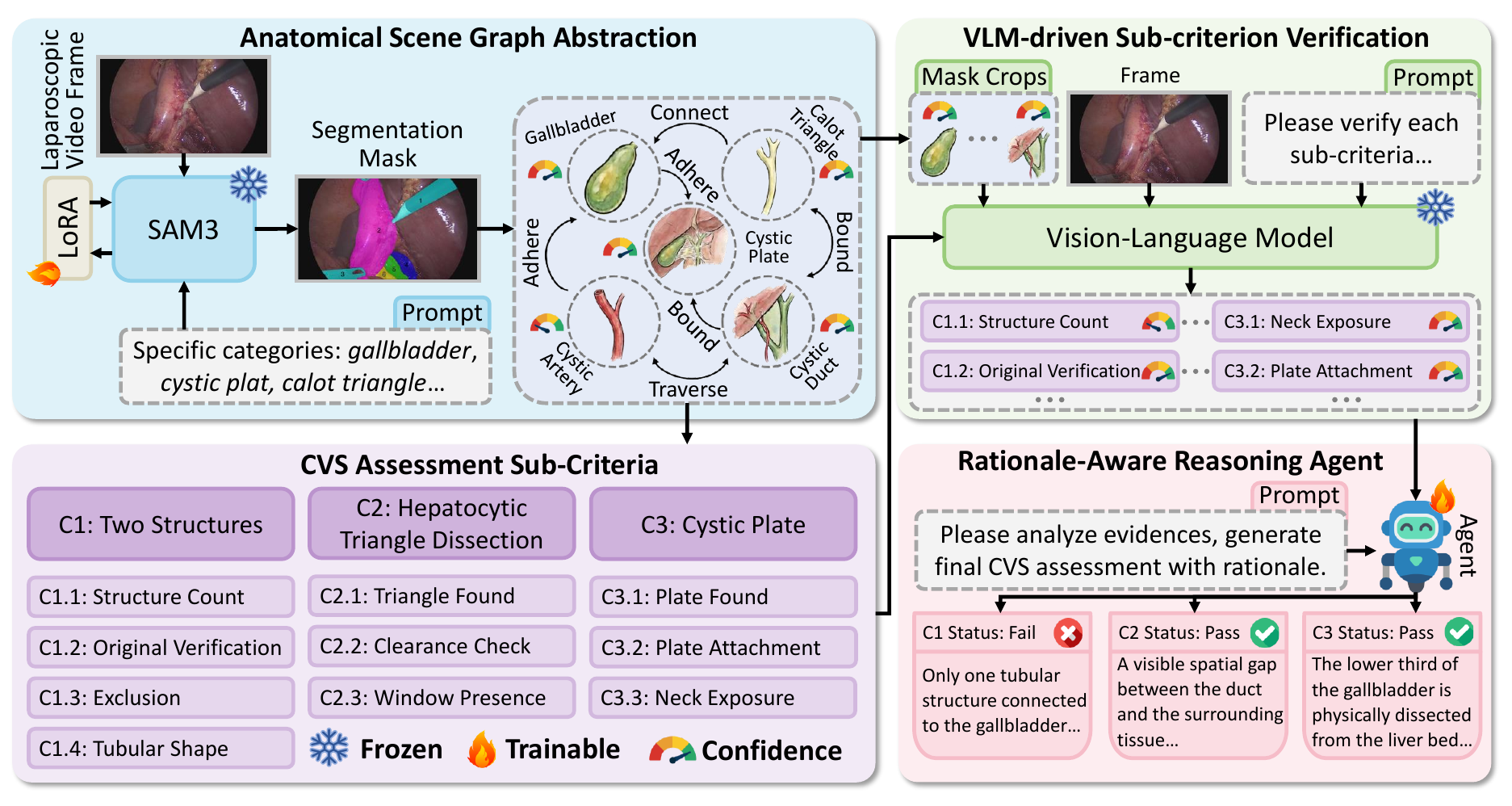}
    \caption{The overview of our ReasonCVS framework for interpretable CVS assessment, consisting of Anatomical Scene Graph Abstraction (ASGA), VLM-driven Sub-criterion Verification, and Rationale-Aware Reasoning Agent. ReasonCVS decomposes CVS assessment into explicit anatomical verification with traceable clinical rationale.}
    \label{fig:method}
\end{figure}

\section{Methodology}

As illustrated in Fig.~\ref{fig:method}, we propose ReasonCVS, a structured reasoning agentic framework that decomposes CVS assessment into explicit verification of fine-grained anatomical relationships. Specifically, we devise an ASGA that organizes anatomical entities and their spatial relationships into a structured representation, VLM-driven Sub-criterion Verification that parses the graph and independently evaluates individual sub-criteria, and a Rationale-Aware Reasoning Agent that synthesizes distributed evidence through calibrated soft reasoning, enabling ReasonCVS to achieve interpretable and accurate CVS assessment.

\subsection{Anatomical Scene Graph Abstraction}
Existing CVS assessment methods \cite{mascagni2022artificial,murali2023latent} typically map visual features directly to criterion-level labels, limiting interpretability and fine-grained anatomical reasoning. However, CVS evaluation is inherently structured around the spatial and relational properties of key anatomical entities. Inspired by scene graph representations in visual understanding \cite{ozsoy20224d,wang2018videos}, we devise the ASGA module that reformulates CVS assessment as semantic graph verification by defining a reference graph $G^{*}$ following \cite{murali2023endoscapes}, where nodes denote expected anatomical structures, edges encode required spatial relationships, and node/edge attributes specify fine-grained constraints.

To construct the observed graph $G_{\text{obs}}$ from the input image $I$, we first employ a semantic segmentation model to obtain per-class anatomical masks, providing the geometric foundation for graph construction. We then perform class-aware instance derivation, where each anatomical class yields a single instance while surgical instruments are separated via connected component analysis. The resulting instances are organized into a structured graph:
\begin{equation}
  G_{\text{obs}} = (V, E), \quad V = \{v_j = (l_j, \sigma_j, m_j)\}_{j=1}^{N}, \quad E = \{e_{jk} = r(v_j, v_k)\}_{j \neq k},
\end{equation}
where each node $v_j$ represents a detected anatomical entity with its semantic label $l_j$, segmentation confidence $\sigma_j$, and binary mask $m_j$, and each edge $e_{jk}$ encodes the spatial relationship $r(\cdot, \cdot)$ between a pair of nodes. To enrich the graph with visual context, we further generate an annotated overlay $O$ by projecting each instance mask onto $I$ with a class-specific color and numeric identifier to preserve global spatial context, along with local crops $\{R_j\}$ extracted from each instance's bounding region to capture fine-grained appearance details. In this way, the proposed ASGA module organizes anatomical entities and their spatial relationships into a structured representation, laying the foundation for explicit, fine-grained anatomical verification.

\subsection{VLM-driven Sub-criterion Verification}
With the structured graph $G_{\text{obs}}$ constructed, we introduce VLM-driven Sub-criterion Verification to parse it into verifiable units and assess each one against the reference graph $G^{*}$. Specifically, we decompose each CVS criterion into a set of fine-grained sub-criteria, where each sub-criterion corresponds to a specific node attribute or edge property in $G^{*}$. This parsing step yields $K$ sub-criteria in total, each associated with a targeted verification prompt $p_i$ that specifies the expected condition to be checked.

For verification, we design criterion-specific verifiers $\{\phi_k\}_{k=1}^{3}$, each dedicated to the sub-criteria of one CVS criterion, allowing it to focus on the relevant anatomical regions and reasoning demands. Each verifier $\phi_k$, powered by a VLM, takes as input the annotated overlay $O$, relevant local crops $\{R_j\}$, and the structured graph, and produces a structured assessment tuple for every sub-criterion $i$:
\begin{equation}
  (d_i, c_i, e_i) = \phi_k(O, \{R_j\}, G_{\text{obs}}, \; p_i),
\end{equation}
where $d_i \in \{0,1\}$ is the binary judgment, $c_i \in [0,1]$ is a confidence score reflecting the reliability of the judgment, and $e_i$ is a natural-language explanation grounding the verdict in the observed anatomical evidence. The complete evidence set $\mathcal{S} = \{(d_i, c_i, e_i)\}_{i=1}^{K}$ is then forwarded to the Rationale-Aware Reasoning Agent for criterion-level synthesis. Through this design, the VLM-driven Sub-criterion Verification serve as specialized perceptual tools that parse the graph and independently evaluate individual sub-criteria, providing interpretable, criterion-level explanations for reliable CVS assessment.

\subsection{Rationale-Aware Reasoning Agent}
Given the distributed observations $\mathcal{S} = \{(d_i, c_i, e_i)\}_{i=1}^{K}$ from the VLM-driven Sub-criterion Verification, we devise a Rationale-Aware Reasoning Agent, fine-tuned via rationale distillation, to synthesize all sub-criterion assessments into per-criterion verdicts with traceable clinical rationales. Functioning as the central decision-maker, the agent takes as input the evidence set $\mathcal{S}$, and produces the final verdicts $\hat{y}$ accompanied by a natural-language rationale $\mathcal{R}$:
\begin{equation}
  \hat{y},\;\mathcal{R} = f_\theta(\mathcal{S}).
\end{equation}
Rather than treating each sub-criterion as an equally weighted binary vote, the agent performs \emph{calibrated soft reasoning}. It cross-examines the collected evidence, upweighting high-confidence assessments while discounting uncertain ones, and resolving potential conflicts across sub-criteria. The generated rationale $\mathcal{R}$ explicitly references specific sub-criterion evidence, allowing clinicians to trace every aspect of the final decision back to inspectable observations. Through this grounded reasoning process, the Rationale-Aware Reasoning Agent serves as an auditable decision-making layer, ensuring that every criterion-level verdict is transparent and traceable to specific anatomical evidence.

\subsection{Optimization Pipeline}
In the design of ReasonCVS, we adopt SAM-3 \cite{carion2025sam} as the segmentation backbone in ASGA, freezing the pre-trained weights and inserting LoRA \cite{hu2022lora} modules into the attention layers for parameter-efficient fine-tuning. The VLM-driven Sub-criterion Verification $\{\phi_k\}_{k=1}^{3}$ are directly prompted without additional training, while the Rationale-Aware Reasoning Agent $f_\theta$ requires fine-tuning to learn effective evidence synthesis. A key challenge is the absence of sub-criterion-level ground truth: the dataset provides only image-level binary labels for each CVS criterion. To address this, we adopt a \emph{rationale distillation} strategy comprising a teacher phase and a student phase.

In the teacher phase, we leverage Gemini-3-pro \cite{team2023gemini} as the teacher model. Given the distributed observations $\mathcal{S}$ together with the ground-truth label $\hat{y}^{*}$, the teacher generates a golden rationale $\mathcal{R}^{*}$, a step-by-step reasoning chain that connects the evidence to the correct verdict, including explicit handling of ambiguous or conflicting observations. This yields a training corpus of $(\mathcal{S}, \mathcal{R}^{*}, \hat{y}^{*})$ triples. In the student phase, we adopt Qwen-3 32B \cite{yang2025qwen3} as the reasoning agent and fine-tune it with LoRA on these triples to generate both the rationale $\mathcal{R}$ and the final verdict $\hat{y}$ conditioned on $\mathcal{S}$. By learning to reproduce the teacher's reasoning patterns rather than merely memorizing label associations, the agent internalizes how to weigh uncertain evidence and resolve conflicts across sub-criteria, yielding robust and interpretable criterion-level judgments.

\begin{table}[!t]
\centering
\caption{Comparison with state-of-the-art methods on Endoscapes-CVS201 for CVS assessment. $\dagger$ denotes spatiotemporal methods. ``-'' indicates that the result is not reported in the original paper. Best results are in \textbf{bold}, second best \underline{underlined}.}
\label{tab:comparison}
\resizebox{\textwidth}{!}{
\begin{tabular}{l cc cccc cccc}
\hline
\multirow{2}{*}{Method} & \multicolumn{2}{c}{Supervision} & \multicolumn{4}{c}{AP (\%)} & \multicolumn{4}{c}{BAcc (\%)} \\
\cline{2-3} \cline{4-7} \cline{8-11}
& CLS & Seg. & Avg & C1 & C2 & C3 & Avg & C1 & C2 & C3 \\
\hline
ResNet50~\cite{murali2023latent} & \multirow{5}{*}{\usym{1F5F8}} &  & 51.5 & 42.9 & 46.5 & 65.1 & 66.7 & 63.5 & 68.1 & 68.4 \\
SurgVLP-vision~\cite{yuan2025learning} &  &  & 51.4 & 51.6 & 41.1 & 61.4 & - & - & - & - \\
HecVL-vision~\cite{yuan2024hecvl} &  &  & 50.0 & 47.8 & 42.4 & 59.7 & - & - & - & - \\
PeskaVLP-vision~\cite{yuan2024procedure} &  &  & 48.9 & 48.3 & 42.4 & 56.2 & - & - & - & - \\
CVS-AdaptNet~\cite{baby2025multi} &  &  & 57.6 & 54.5 & 55.9 & 62.4 & - & - & - & - \\
\hline
LayoutCVS~\cite{murali2023latent} & \multirow{6}{*}{\usym{1F5F8}} & \multirow{6}{*}{\usym{1F5F8}} & 57.1 & 62.2 & 50.5 & 58.5 & 69.4 & 68.2 & 70.8 & 69.2 \\
ResNet50-DetInit~\cite{murali2023latent} &  &  & 58.5 & 58.5 & 54.9 & 62.0 & 70.5 & 72.2 & 70.0 & 69.2 \\
LG-CVS~\cite{murali2023latent} &  &  & 63.2 & 67.9 & 53.6 & 68.0 & \underline{74.8} & 75.8 & \underline{75.4} & 73.2 \\
DeepCVS$^\dagger$~\cite{mascagni2022artificial} &  &  & 60.5 & 63.5 & 53.8 & 64.4 & 73.5 & 75.6 & 73.8 & 71.1 \\
STRG$^\dagger$~\cite{wang2018videos} &  &  & 59.0 & 52.4 & 55.5 & 69.2 & 68.4 & 62.3 & 69.1 & \underline{73.7} \\
SV2LSTG$^\dagger$~\cite{murali2023encoding} &  &  & \underline{64.3} & \underline{66.0} & \underline{56.4} & \underline{70.6} & 73.4 & \underline{76.2} & 73.8 & 70.3 \\
\hline
ReasonCVS (Ours) & \usym{1F5F8} & \usym{1F5F8} & \textbf{68.1} & \textbf{67.5} & \textbf{62.9} & \textbf{73.7} & \textbf{81.4} & \textbf{79.2} & \textbf{83.7 }& \textbf{81.4} \\
\hline
\end{tabular}
}
\end{table}

\section{Experiments}

\subsection{Experimental Setup}
To validate the effectiveness of the proposed ReasonCVS, we conduct experiments on the Endoscapes-CVS201 benchmark \cite{mascagni2025endoscapes} for CVS assessment, which evaluates three CVS criteria: C1 (Two Structures), C2 (Hepatocystic Triangle Dissection), and C3 (Cystic Plate). The dataset comprises 11,090 frames with image-level CVS annotations, and 493 frames with pixel-level segmentation masks. We follow the official split for training, validation, and test sets for both tasks. All experiments are performed on a single NVIDIA H200 GPU using PyTorch. For fine-tuning SAM-3 in ASGA, we employ the AdamW optimizer with a learning rate of $1\times10^{-4}$ and adopt CosineAnnealingLR as the scheduling strategy. For fine-tuning the Rationale-Aware Reasoning Agent, we adopt Qwen-3 32B as the backbone and apply LoRA with rank 64 and alpha 128. We use the AdamW optimizer with a learning rate of $2\times10^{-5}$, and train for 3 epochs. For quantitative evaluation, we adopt mean Average Precision (mAP) to measure ranking quality across confidence thresholds, and Balanced Accuracy (BAcc) to account for class imbalance by averaging per-class accuracy.

\begin{figure}
    \centering
    \includegraphics[width=0.9\linewidth]{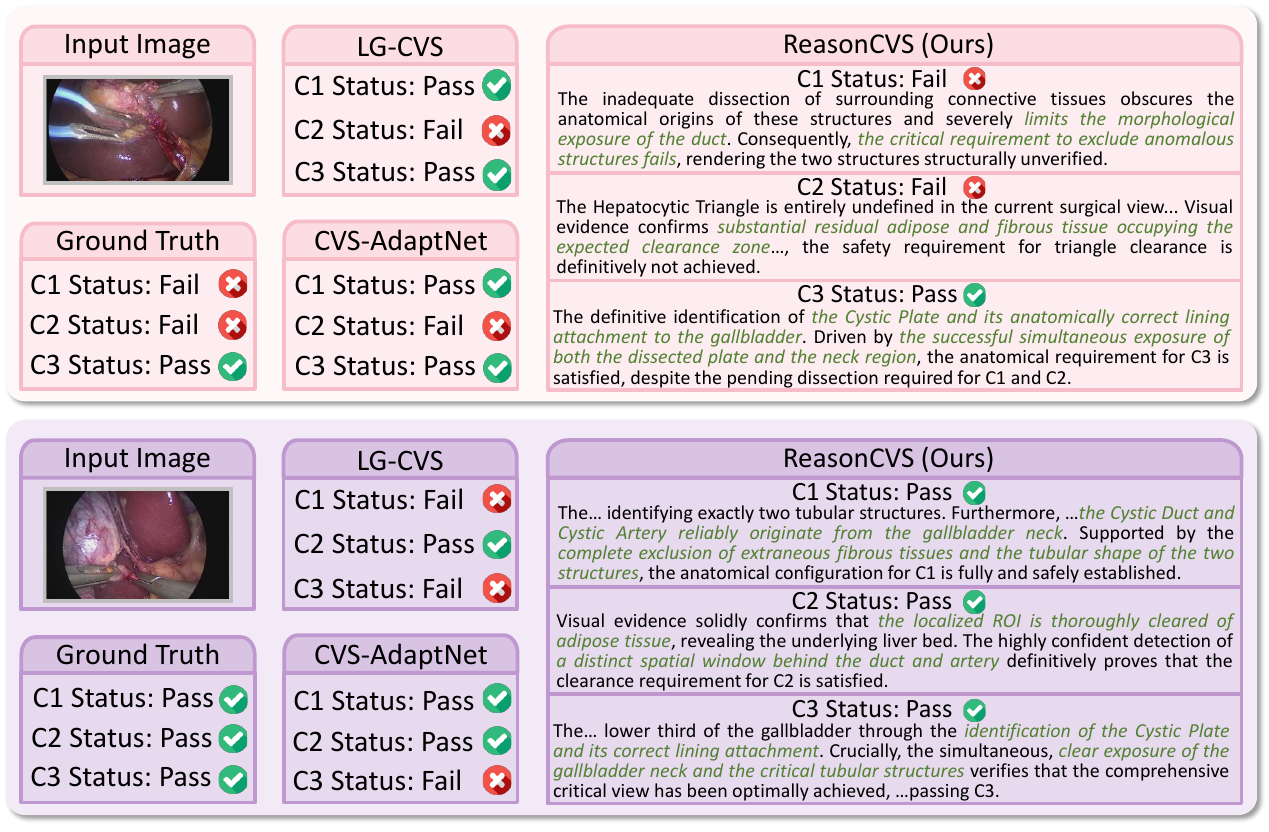}
    \caption{Qualitative comparison of CVS assessment results. Our ReasonCVS provides correct, criterion-level rationales grounded in specific anatomical evidence, enabling interpretable and traceable clinical reasoning.}
    \label{fig:visualization}
\end{figure}

\begin{table}[!t]
\centering
\caption{Ablation study of our CVSAgent on Endoscapes-CVS201 for CVS
assessment.}
\tabcolsep=0.18cm
\label{tab:ablation_components}
\resizebox{\textwidth}{!}{
\begin{tabular}{ccc cccc cccc}
\hline
\multirow{2}{*}{ASGA} & \multirow{2}{*}{VSV} & \multirow{2}{*}{RARA} & \multicolumn{4}{c}{mAP (\%)} & \multicolumn{4}{c}{BAcc (\%)} \\
\cline{4-7} \cline{8-11}
& & & Avg & C1 & C2 & C3 & Avg & C1 & C2 & C3 \\
\hline
&  &  & 43.6 & 41.7 & 45.2 & 43.9 & 56.5 & 61.2 & 54.9 & 53.5 \\
\usym{1F5F8} &  &  & 54.2 & 53.5 & 52.8 & 56.2 & 64.8 & 66.7 & 63.4 & 64.3 \\
\usym{1F5F8} & \usym{1F5F8} &  & 60.3 & 59.1 & 57.6 & 64.3 & 72.4 & 72.4 & 73.1 & 71.6 \\
\usym{1F5F8} & \usym{1F5F8} & \usym{1F5F8} & \textbf{68.1} & \textbf{67.5} & \textbf{62.9} & \textbf{73.7} & \textbf{81.4} & \textbf{79.2} & \textbf{83.7} & \textbf{81.3} \\
\hline
\end{tabular}
}
\end{table}

\subsection{Comparison with State-of-the-art Methods}
As shown in Table~\ref{tab:comparison}, image-only methods~\cite{murali2023latent,yuan2024procedure,yuan2024hecvl,yuan2025learning} achieve limited mAP (48.9--51.5\%) due to the lack of explicit anatomical reasoning. CVS-AdaptNet~\cite{baby2025multi} improves to 57.6\% via vision-language pretraining but still lacks structural understanding. Segmentation-based methods such as LG-CVS~\cite{murali2023latent} (63.2\%) and the temporal approach SV2LSTG~\cite{murali2023encoding} (64.3\%) leverage segmentation for spatial reasoning, yet all treat CVS assessment as a holistic prediction without sub-criterion verification. Our ReasonCVS achieves the best performance with 68.1\% mAP, surpassing SV2LSTG by 3.8\%. The largest gain appears on C2 (62.9\% vs.\ 56.4\%), where complex hepatocystic triangle dissection particularly benefits from structured graph-based reasoning. ReasonCVS also attains 81.4\% BAcc (+6.6\% over LG-CVS), confirming its robustness under class imbalance.

\subsection{Ablation Study}

To validate the effectiveness of our proposed ASGA, VLM-driven Sub-criterion Verification (VSV), and Rationale-Aware Reasoning Agent (RARA), we present the results of ablation studies in Table~\ref{tab:ablation_components}. By introducing ASGA, the model achieves 10.6\% mAP and 8.3\% BAcc improvements. Incorporating VSV yields additional gains of 6.1\% mAP and 7.6\% BAcc, confirming that fine-grained sub-criterion decomposition enables more precise anatomical assessment. Notably, integrating RARA brings the most significant improvements of 7.8\% mAP and 9.0\% BAcc, demonstrating that rationale distillation effectively teaches the agent to weigh uncertain evidence and resolve conflicts across sub-criteria. These results validate that ASGA, VSV, and RARA collectively contribute to the superior performance of ReasonCVS for interpretable CVS assessment.

\begin{figure}[!t]
    \centering
    \includegraphics[width=0.8\linewidth]{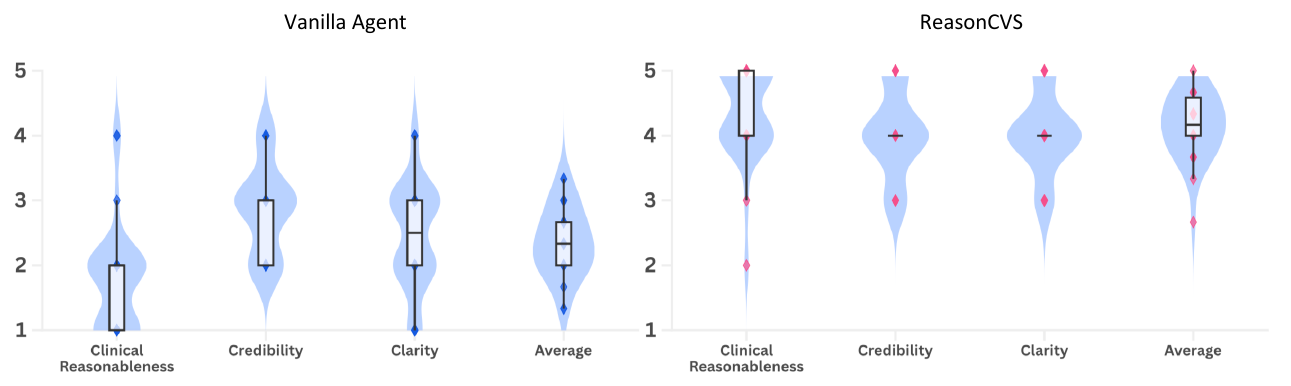}
    \caption{Distribution of expert evaluation scores for the Vanilla Agent and ReasonCVS across four dimensions. ReasonCVS consistently achieves higher and more concentrated scores, indicating superior and more stable reasoning quality.}
    \label{fig:violin}
\end{figure}

\subsection{Clinical Evaluation of Reasoning Quality}

To validate clinical interpretability, a certified surgeon scores the reasoning outputs of 30 samples (10 per criterion) on a 1--5 Likert scale across three dimensions: \textit{Clinical Reasonableness} (alignment with surgical logic), \textit{Credibility} (soundness of argumentation), and \textit{Clarity} (ease of interpretation). A Vanilla Agent (same Qwen-3 32B backbone, identical inputs, no rationale distillation) serves as the reference. As shown in Fig.~\ref{fig:violin}, ReasonCVS yields consistently higher and more concentrated distributions across all dimensions, whereas the Vanilla Agent spreads widely toward the lower end. The contrast is most striking on Clinical Reasonableness (ReasonCVS centers at 4--5 vs.\ 1--2 for Vanilla), confirming that rationale distillation is essential for clinically grounded reasoning.

\section{Conclusion}
In this work, we present ReasonCVS, a structured reasoning agentic framework empowered by Vision-Language Models for interpretable Critical View of Safety assessment. We devise an Anatomical Scene Graph Abstraction that organizes anatomical entities and their spatial relationships into a structured representation. To operationalize this, we introduce VLM-driven Sub-criterion Verification that parses the graph and independently evaluates individual sub-criteria, along with a Rationale-Aware Reasoning Agent fine-tuned via rationale distillation that synthesizes distributed observations through calibrated soft reasoning, yielding a final verdict alongside a traceable clinical rationale. Extensive experiments on the Endoscapes-CVS201 benchmark demonstrate that ReasonCVS achieves superior performance (68.1\% mAP) over state-of-the-art methods while providing interpretable, criterion-level explanations for reliable surgical assessment.

\bibliographystyle{splncs04}
\bibliography{refs}
\end{document}